\documentclass[runningheads]{llncs}

\usepackage{wrapfig,booktabs}
\usepackage{graphicx}

\usepackage{tikz}
\usepackage{comment}
\usepackage{amsmath,amssymb} 
\usepackage{color}

\usepackage{subfigure}
\usepackage{algpseudocode}
\usepackage{algorithm}
\usepackage{hyperref}
\hypersetup{
    colorlinks=true,
    linkcolor=blue,
    filecolor=magenta,      
    urlcolor=cyan,
}

\begin{document}
\pagestyle{headings}
\mainmatter
\def\ECCVSubNumber{18}  

\title{Bayesian Zero-Shot Learning} 

\titlerunning{Bayesian Zero-Shot Learning}
%
\author{Sarkhan Badirli\inst{1} \and
Zeynep Akata\inst{2} \and
Murat Dundar\inst{3}}
\authorrunning{S. Badirli et al.}
%
\institute{Purdue University, Computer Science Department, USA \and University of T\"ubingen, Cluster of Excellence Machine Learning, Germany
\and
IUPUI, Computer and Information Science Department, USA}
\maketitle

\begin{abstract}
Object classes that surround us have a natural tendency to emerge at varying levels of abstraction. We propose a Bayesian approach to zero-shot learning (ZSL) that introduces the notion of meta-classes and implements a Bayesian hierarchy around these classes to effectively blend data likelihood with local and global priors. Local priors driven by data from seen classes, i.e., classes available at training time, become instrumental in recovering unseen classes, i.e., classes that are missing at training time, in a generalized ZSL (GZSL) setting. Hyperparameters of the Bayesian model offer a convenient way to optimize the trade-off between seen and unseen class accuracy. We conduct experiments on seven benchmark datasets, including a large scale ImageNet and show that our model produces promising results in the challenging GZSL setting.
\end{abstract}

\section{Introduction}

Natural images exhibit power-law property; hence, in a randomly sampled training set, no training examples are expected to be available for most of the object categories \cite{imagenet,dl1,dl2}. This restriction becomes more evident in a fine-grained object recognition task.  
Zero-shot learning (ZSL), which considers training and test classes, i.e. seen and unseen classes, as two disjoint sets, was introduced to mitigate this limitation \cite{lampert_cvpr,palatucci2009zero}. The two groups of classes are linked through a shared set of attributes that characterize high level semantic descriptions of all classes. During the training phase, a mapping between examples of seen classes and their corresponding class-based attributes is learned. This mapping is later used to identify examples of unseen classes during the test phase. 

The standard ZSL setting restricts test time search space to only unseen classes. This somewhat unrealistic stipulation was later relaxed in the generalized ZSL (GZSL) setting to include all classes during the test phase~\cite{gzsl}.   
In GZSL, side information, i.e., attributes, are as important as the perceptual representation of images. Attribute vectors are either manually annotated~\cite{lampert_cvpr,farhadi} or derived from free-form text using word embedding~\cite{cmt,devise,w2v}.  
Early line of work in ZSL~\cite{lampert_cvpr} assumes attribute independence and uses probabilistic classifiers to assign images to test classes.  

In this paper, we tackle ZSL by introducing a two layer Bayesian hierarchy  manifesting over both seen and unseen classes.  Our approach is designed to leverage the implicit hierarchy present among classes, especially evident in fine grained data sets~\cite{cub,flo,sun}. Unlike earlier approaches, which seek to optimize an embedding between image and semantic spaces, the proposed method assumes that there are latent classes that define the class hierarchy in image space and uses semantic information to build the Bayesian hierarchy around these meta-classes.  

\begin{wrapfigure}{r}{0.50\textwidth}
    \centering
    \vspace{-20pt}
    \includegraphics[width=0.5\textwidth]{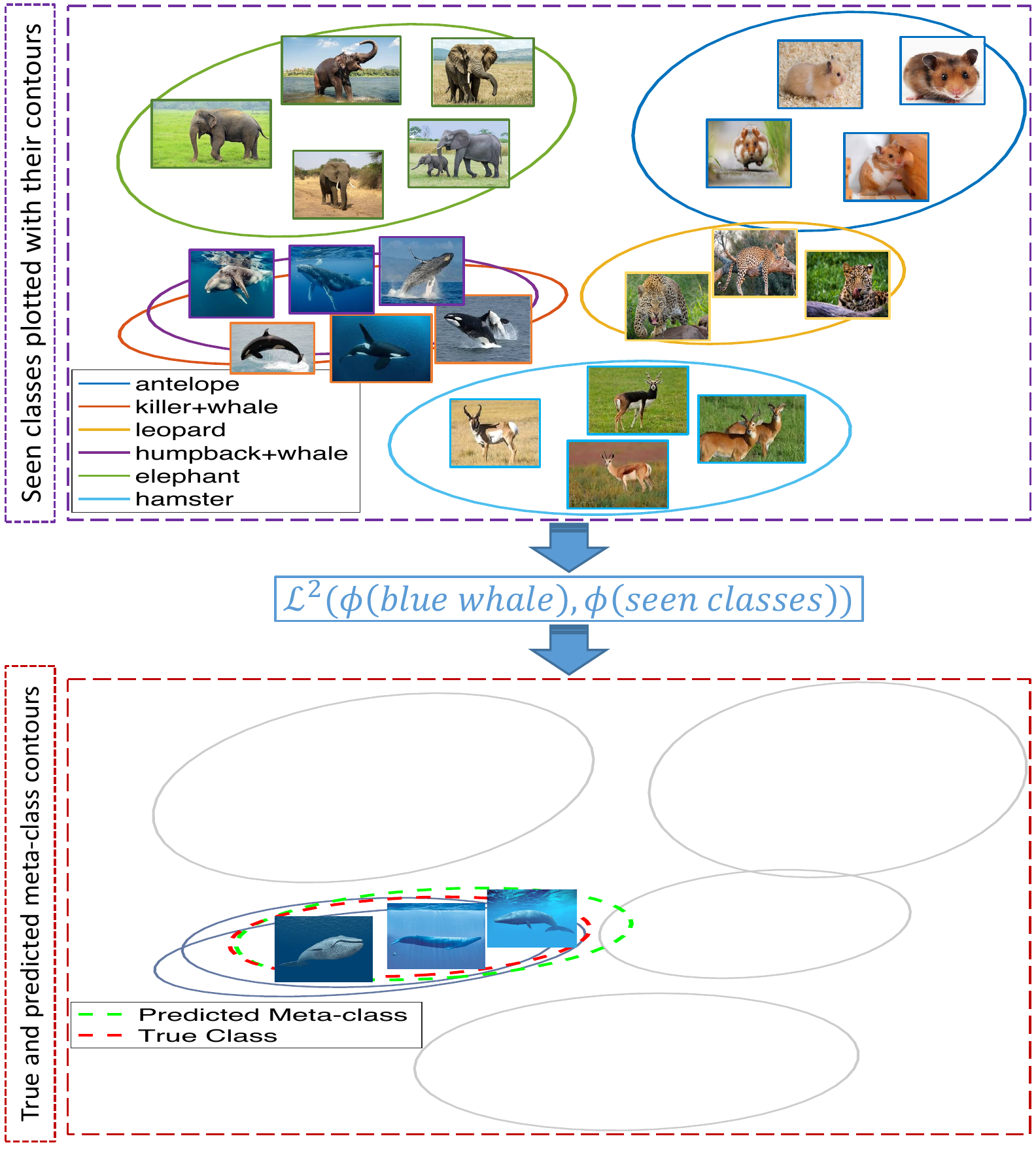}
    \caption{Meta-classes illustrated in 2D PCA space (reduced from 2048).  Only a small subset of seen classes are shown. Contours are derived from class covariance matrices and  placed at two standard deviations away from class means. Meta class for \textit{blue whale} (unseen) predicted based on \textit{killer} and \textit{humpback} whales (seen).}
    \vspace{-10pt}
    \label{fig:process}
\end{wrapfigure}

Our model uses two types of Bayesian priors: global and local. As the name suggests, global priors are shared across all classes, whereas local priors are only shared among semantically similar classes, which are identified based on the distances between attribute vectors in the Euclidean space. Unlike standard Bayesian models where the posterior predictive distribution establishes a compromise between prior and likelihood, our approach utilizes posterior predictive distributions to reconcile information about local and global priors as well as the likelihood to more effectively accommodate the class hierarchy. In this framework, unseen classes are represented by their corresponding meta classes (see Figure~\ref{fig:process}), and test samples are classified based on posterior predictive likelihoods computed for both seen and unseen classes. Our approach achieves significant improvements on both seen and unseen class accuracies to achieve the best results on a variety of benchmark datasets among the currently published state of the art methods.  

Our contributions are as follows.
(1) We propose a hierarchical Bayesian model based on the intuition that actual classes originate from their corresponding local priors, each defined by a meta-class of its own. 
(2) We derive the posterior predictive distribution (PPD) for a two-layer Gaussian mixture model to effectively blend local and global priors with data likelihood. These PPDs are used to implement a maximum-likelihood classifier, which represents seen classes by their own PPDs and unseen classes by meta-class PPDs. 
(3) Across seven datasets with varying granularity and sizes, in particular on the large-scale ImageNet dataset, we show that the proposed model is highly competitive against existing inductive techniques in the GZSL setting.

\section{Related Work}

In this section, we discuss the prior work on zero-shot learning and hierarchical generative models related to ours.

\textbf{Zero-Shot Learning.} There has been an increasing interest in classifying fine grained and large-scale image datasets~\cite{sun,cub,flo,imagenet}. This, in turn, led to a surge of interest in ZSL as labeling them is extremely costly. In their seminal paper~\cite{lampert_cvpr}, authors tackle ZSL by implementing a probabilistic classifier for each attribute and then classifying test cases by aggregating attribute probabilities for each class. This approach treats attributes as independent, which is a fairly strong assumption for most real-world data sets. This work was followed by a large body of work that seeks to optimize a mapping from image space, i.e., feature vectors, onto semantic space, i.e., attribute vectors. This line of work can be categorized into two according to whether the mapping is bi-linear~\cite{devise,sae,eszsl,ale,sje}  or non-linear~\cite{cmt,latem}.  Related to ours, \cite{sse,conse,sync} first maps image and semantic space into an intermediate space and represents unseen classes as a mixture of seen classes. Besides these mainline ZSL studies, a recent study evaluates an extended version of a few-shot learning algorithm for ZSL \cite{Rel_net}. This approach learns a deep metric to query images with few shot samples. Extension to ZSL is achieved by replacing few-shot samples with one-shot class attribute vectors. 

\textbf{Generative models for ZSL.} Although most of the early work focused on discriminative models, there are a few studies that use generative models to tackle ZSL \cite{gen1,gfzsl}. The study in \cite{gen1} uses Normal distributions to model both image features and semantic vectors and learns a multimodal mapping between two spaces. This mapping is optimized by minimizing a  similarity based cross domain loss function. In a similar fashion the study in \cite{gfzsl} utilizes a regression model to optimize a mapping between class attributes and parameters of class conditional distributions. A comprehensive review of these techniques and their performance on several benchmark data sets can be found in \cite{gbu_tpami}. 

There are also quite a few techniques that tackle ZSL in a transductive setting. Experiments in \cite{f_clsgan,cada_vae} demonstrate that unlabeled data from unseen classes as well as training data augmented by generative adversarial nets/ variational autoencoders can notably boost the classification accuracy. We believe that this line of work should be treated under a different category as a direct comparison with current ZSL techniques is not possible since similar data augmentation techniques could have most certainly benefited these techniques.

\textbf{Bayesian models.} In this paper, we offer a hierarchical Bayesian perspective on ZSL as a promising alternative to earlier approaches. Although hierarchical Bayesian mixture models have been previously explored for a variety of clustering problems \cite{hdpre,aspire_aml,aspire_rare,i2gmm}, their extension to ZSL comes with two distinct features that could help the proposed model prevail over the large body of early work in ZSL. First, as a Bayesian model, ours offers a systematic approach to sharing information between seen and unseen classes as well as within each group through the utilization of local and global priors. Global priors are defined by hyperparameters, whereas local priors are determined by the parameters of the meta classes, which are estimated from corresponding seen classes. Second, as a hierarchical model, it can better accommodate data sets with different levels of class abstractions, i.e., fine-grained vs. coarse-grained data sets, which is particularly appealing for large-scale classification. A hierarchical Bayesian model was previously studied in a one-shot learning setting \cite{salakhutdinov}. Our proposed approach differs from this model in two essential aspects. First, unlike our proposed approach, no semantic information was used when establishing the Bayesian hierarchy in \cite{salakhutdinov}, and class discovery was performed in a fully unsupervised fashion. Second, our approach introduces the notion of local prior, which becomes highly instrumental in defining meta-classes and modeling dispersion of classes.

\begin{figure*}[t]
  \centering
  \subfigure[Our Bayesian Zero-Shot Learning Model (BZSL)]{\includegraphics[width=0.65\textwidth]{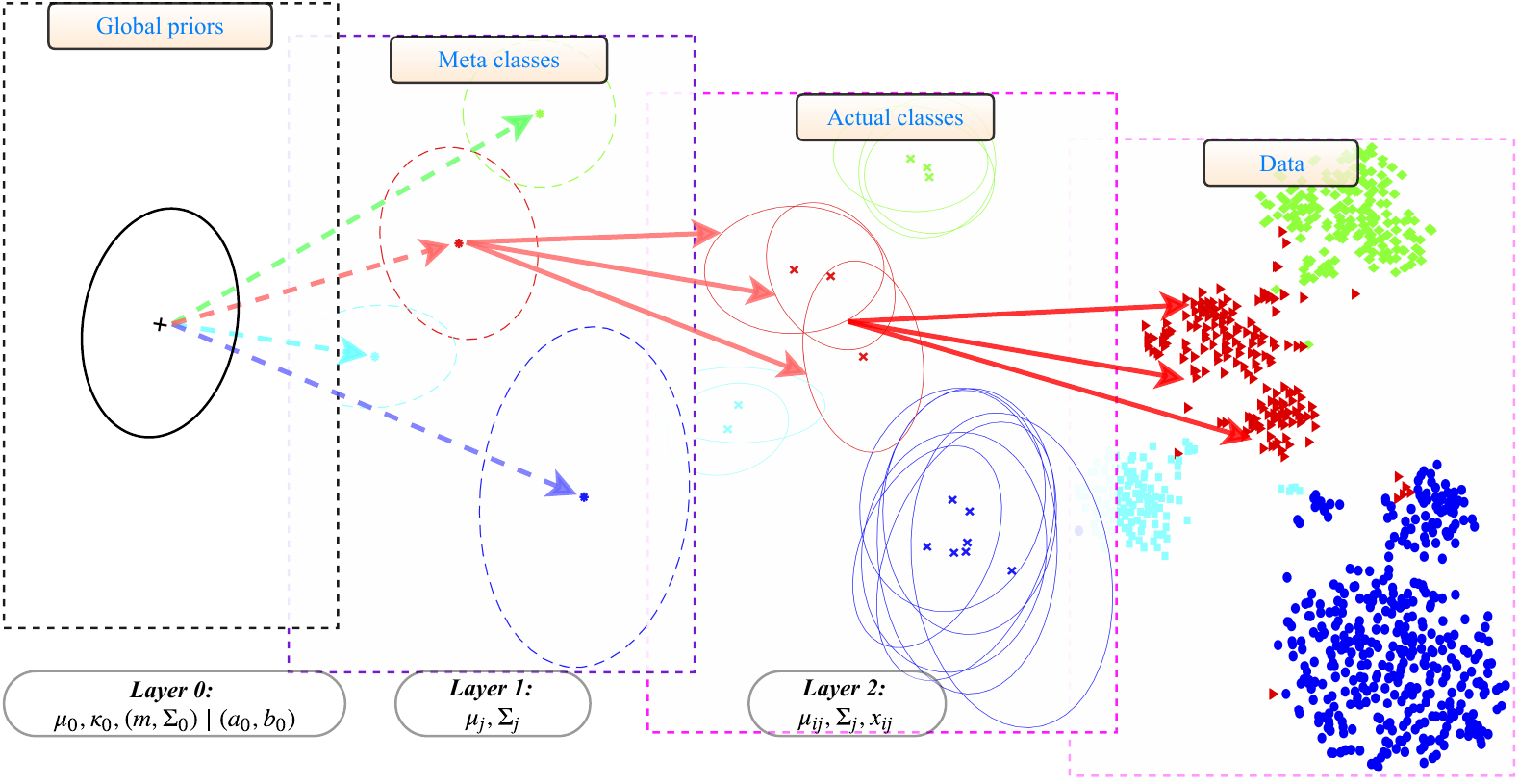}} \hfill
  \subfigure[Graphical model]{\includegraphics[width=0.33\textwidth]{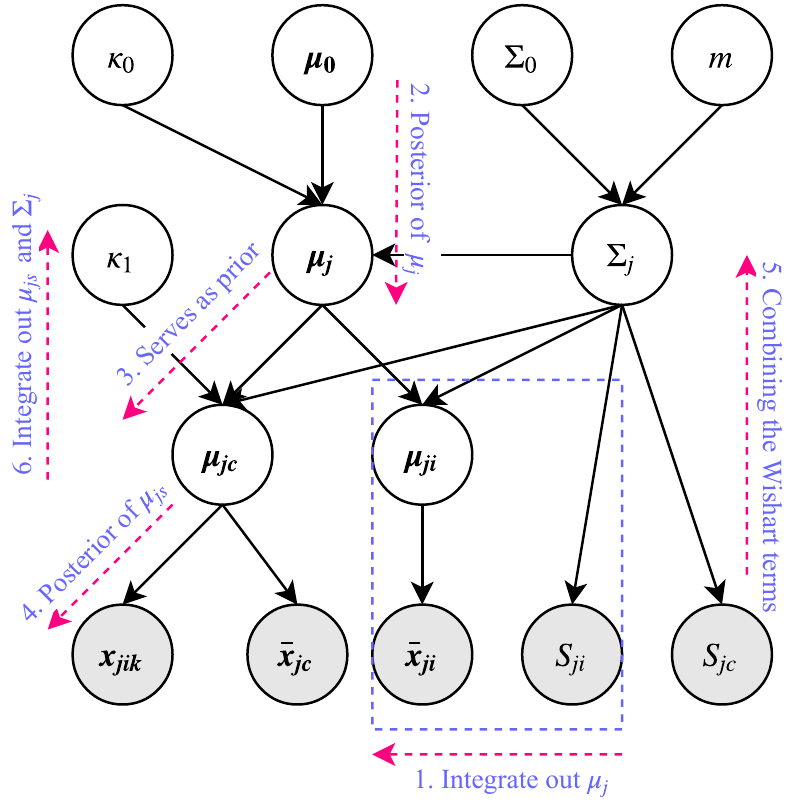} \label{fig:graphical}}
  \caption{Two-layer Generative Model for Bayesian zero-shot learning (BZSL). (a) Latent (Meta) classes shown by dashed lines in layer 1 are generated from Normal distribution ($N(\boldsymbol{\mu_{j}}, \Sigma_{j})$). Hyperparameters in layer 0 are the priors on the mean and covariance of these Gaussians. The sufficient statistics of actual classes in layer 2 are obtained from layer 1. Finally data samples are drawn from Gaussian distribution with the mean, $\boldsymbol{\mu_{ji}}$ and covariance, $\Sigma_{j}$. (b) Conditional hierarchical Gaussian data generation (Likelihood) model and derivation of marginal likelihood.}
  \label{fig:diagrams}
\end{figure*}

\textbf{Our work.} Unlike the vast majority of early work, which seeks to optimize a mapping between image features and attribute vectors, our approach readily models class distributions in the feature space by exploiting both local and global priors defined over the parameters of these distributions. Local priors are defined by meta-classes. In the proposed approach, attribute vectors only come into play when determining meta-class memberships of actual classes. Classes with similar attribute vectors are pooled together to derive local priors. 

\section{Bayesian Zero Shot Learning}

Bayesian classification places a shared prior over the parameters of class distributions, which are assumed to be generated independently conditioned on the prior. Imposing the same prior across all classes creates dependencies among them, enabling information propagation and regularization at the same time during model inference. However, in real-world applications the classes are often not generated independently, indeed for a large number of classes different levels of abstraction is expected. On the other hand, availability of semantic side information suggests that there is a deeper level of hierarchy among existing classes than a single global Bayesian prior can explain. 

Images from semantically similar classes are embedded close to each other due to their shared latent parameters. When such similarities are not accounted for in the classification model, sample estimates of class parameters derived based on independence assumption among classes become nullified. In other words, knowing the parameters of the global prior may not be sufficient for achieving independence as latent parameters define deeper level hierarchical relationships among classes. Our model resolves this problem by introducing a layer of meta-classes between global prior and actual classes, paving the way for independence and enabling information sharing and propagation across classes.

\subsection{Generative Model}

Our approach to ZSL employs class similarities by a two layer generative model. As shown in Figure~\ref{fig:diagrams}, our model identifies meta-classes that determine groupings among classes. These meta-classes play a key role by acting as a local prior for individual classes, i.e. both seen and unseen classes that belong to the same meta-class inheriting the same local prior. In our framework, the data points with the same local prior that do not belong to any of the seen classes can be considered from unseen classes. If the class groupings can be arranged such that there is only one unseen class associated with each local prior then unseen classes can be uniquely identified. Associating each unseen class with a different local prior forms the basis of our approach. Our generative model is designed as follows: 
\begin{gather}
\resizebox{\linewidth}{!}{%
$\boldsymbol{x_{jik}}  \sim   N(\boldsymbol{\mu_{ji}},\Sigma_{j}), \quad
\boldsymbol{\mu_{ji}}  \sim  N(\boldsymbol{\boldsymbol{\mu_{j}}},\Sigma_{j}\kappa_{1}^{-1}), \quad
\boldsymbol{\mu_{j}}  \sim  N(\boldsymbol{\mu_{0}},\Sigma_{j}\kappa_{0}^{-1}), \quad
\Sigma_{j} \sim W^{-1}(\Sigma_{0},m)$
\label{eq:i2gmm}
}
\end{gather}
with the meta-class index $j$, the actual class index $i$, the image index $k$. We assume that images $\boldsymbol{x_{jik}}$ come from a Gaussian with mean $\boldsymbol{\mu_{ji}}$ and covariance matrix $\Sigma_{j}$. They are generated independently conditioned not only on the global prior but also on their corresponding meta-class. 

Each meta-class is characterized by the parameters $\boldsymbol{\mu_{j}}$ and $\Sigma_{j}$.
$\boldsymbol{\mu_{0}}$ is the mean of the Gaussian prior defined over the mean vectors of meta-classes, $\kappa_{0}$ is a scaling constant that adjusts the dispersion of the centers of meta classes around $\boldsymbol{\mu_{0}}$. A smaller value for $\kappa_{0}$ suggests that class centers are expected to be farther apart from each other whereas a larger value suggests they are expected to be closer to each other. On the other hand, $\Sigma_{0}$ and $m$ dictate the expected shape of the class distributions, as under the inverse Wishart distribution assumption the expected covariance is $E(\Sigma|\Sigma_{0},m)=\frac{\Sigma_{0}}{m-D-1}$, where $D$ is the dimension of image feature space. The minimum feasible value of $m$ is equal to $D+2$, and the larger the $m$ is the less individual covariance matrices will deviate from the expected shape. 

On the other hand, $\kappa_{1}$ is a scaling constant that adjusts the dispersion of the actual class means around their corresponding meta-class means. A larger $\kappa_{1}$ leads to smaller variations in class means compared to the mean of their corresponding meta classes, suggesting a fine-grained relationship among classes sharing the same meta-class. On the other hand, a smaller $\kappa_{1}$ dictates coarse-grained relationships among classes sharing the same meta-class. In this model, classes with the same meta-class also share the same covariance matrix $\Sigma_{j}$ to preserve conjugacy of the model.  

To classify test examples, we need the posterior predictive distributions (PPD) of seen and unseen classes which we will explain next. More details about the derivation are provided in the supplementary.

\subsection{Posterior Predictive Distribution}

In our model, the posterior predictive distribution (PPD) incorporates three sources of information: the data likelihood that arises from the current class, the local prior that results from other classes sharing the same meta class as the current class, and global prior defined in terms of hyperparameters. The derivation in six steps are outlined in Figure~\ref{fig:graphical} and Algorithm\footnote{The code is publicly available at \href{https://github.com/sbadirli/Bayesian-ZSL}{GitHub}} \ref{algo:main} describes a pseudo code on deriving PPD for both seen and unseen classes. Class sufficient statistics are summarized by $\boldsymbol{\bar{x}_{ji}}, S_{ji}$ and $n_{ji}$ which represent sample mean, scatter matrix and size of class $i$ of meta-class $j$, respectively. The notations $\omega_{jc}$ and $\omega_j$   used in the Algorithm \ref{algo:main} represents the current seen class and unseen class, whose PPD is being derived.

In step 1, we establish the link between class sample mean $\boldsymbol{\bar{x}_{ji}}$ and its corresponding meta-class mean $\boldsymbol{\mu_{j}}$ by marginalizing out the intermediate class mean $\boldsymbol{\mu_{ji}}$. As all of these are Gaussians, this marginalization yields a Gaussian:
\begin{eqnarray}
P(\boldsymbol{\bar{x}_{ji}}|\boldsymbol{\mu_{j}},\Sigma_{j},\kappa_{1}) & = & N(\boldsymbol{\bar{x}_{ji}}|\boldsymbol{\mu_{j}},\Sigma_{j}(\frac{1}{n_{ji}}+\frac{1}{\kappa_{1}})) \label{eq:connectocluster}
\end{eqnarray}
In step 2, we use Bayes rule to obtain the posterior distribution of the meta-class mean vector $\boldsymbol{\mu_j}$: 
\begin{gather}
P(\boldsymbol{\mu_{j}}|\boldsymbol{\mu_{0}},\Sigma_{j},\kappa_{0}, \kappa_{1},\{\boldsymbol{\bar{x}_{ji}}\}_{t_i=j}) =  N(\boldsymbol{\mu_{j}}|\boldsymbol{\bar{\mu}_{j}},\bar{\kappa}_{j}^{-1}\Sigma_{j})\nonumber\\
\boldsymbol{\bar{\mu}_{j}}  =  \frac{\sum_{i:t_{i}=j}\frac{n_{ji}\kappa_{1}}{(n_{ji}+\kappa_{1})}\boldsymbol{\bar{x}_{ji}}+\kappa_{0}\boldsymbol{\mu_{0}}}{\sum_{i:t_{i}=j}\frac{n_{ji}\kappa_{1}}{(n_{ji}+\kappa_{1})}+\kappa_{0}}, \quad
\bar{\kappa}_{j} = (\sum_{i:t_{i}=j}\frac{n_{ji}\kappa_{1}}{(n_{ji}+\kappa_{1})}+\kappa_{0}) \label{eq:mj_pd}
\end{gather}
where $t_i$ is the  meta-class indicator for class $i$.  Note that the mean $\boldsymbol{\bar{\mu}_j}$ is the weighted average of the prior mean and class means share the same meta-class.

\begin{algorithm}[t]
\textbf{Input:} Training data, $\phi(seen)$, $\phi(unseen)$\\
\textbf{Output:} PPD parameters for each seen class ($\boldsymbol{\bar{\mu}_{jc}}, \bar{v}_{jc}, \bar{\Sigma}_{jc} $) and unseen class ($\boldsymbol{\bar{\mu}_{j}}, \bar{v}_{j}, \bar{\Sigma}_{j} $) 
\begin{algorithmic}[1]
\State Set hyper-parameters: $\kappa_0, \kappa_1, m, s, K$
\State Compute $\boldsymbol{\mu_0}$ (mean of class means) and $\Sigma_0$ (mean of class covariances scaled by s)
\For{each seen class $\omega_{jc}$}\Comment{Images available}
        \State Calculate current class params: $\boldsymbol{\bar{x}_{jc}}, n_{jc}, S_{jc}$
        \State Find K most similar seen classes: 
        \State $\mathcal{L}^2(\phi(\omega_{jc}),\phi(seen))$
        \For{ each selected seen class $\omega_{ji}$}
                    \State Calculate class params: $\boldsymbol{\bar{x}_{ji}}, n_{ji}, S_{ji}$ 
        \EndFor
        \State Calculate intermediate terms: $\tilde{\kappa}_j, \boldsymbol{\bar{\mu}_j}, S_{\boldsymbol{\mu}}$ (Eq \ref{eq:mu_ks},\ref{eq:mj_pd},\ref{eq:iw}) 
        \State Calculate PPD parameters by combining \textit{local prior} 
        \State and \textit{data driven likelihood}: $\boldsymbol{\bar{\mu}_{jc}}, \bar{v}_{jc}, \bar{\Sigma}_{jc}$ (Eq \ref{eq:i2gmm_seen})
\EndFor 
\For{each unseen class $\omega_{j}$}\Comment{No image available}
        \State Find K most similar seen classes: 
        \State $\mathcal{L}^2(\phi(\omega_{j}),\phi(seen))$
        \For{ each selected seen class $\omega_{ji}$}
                    \State Calculate class params: $\boldsymbol{\bar{x}_{ji}}, n_{ji}, S_{ji}$ 
        \EndFor
        \State Calculate intermediate terms: $\tilde{\kappa}_j, S_{\boldsymbol{\mu}}$ (Eq \ref{eq:mu_ks}, \ref{eq:iw}) 
        \State Calculate PPD parameters using only \textit{local} 
        \State \textit{prior}: $\boldsymbol{\bar{\mu}_{j}}, \bar{v}_{j}, \bar{\Sigma}_{j}$ (Eq \ref{eq:mj_pd}, \ref{eq:i2gmm_seen})
\EndFor 
\end{algorithmic}
 \caption{Modeling seen and unseen classes in BZSL}
 \label{algo:main}
\end{algorithm}

In step 3, we obtain the local prior for class mean vector $\boldsymbol{\mu_{jc}}$ by propagating the information from other classes sharing the same meta-class as the current class $c$. This is achieved by integrating out the meta-class mean vector $\boldsymbol{\mu_{j}}$.
\begin{equation}
P(\boldsymbol{\mu_{jc}}|\boldsymbol{\mu_{0}},\Sigma_{j},\kappa_{0}, \kappa_{1},\{\boldsymbol{\bar{x}_{ji}}\}_{t_i=j}) = N(\boldsymbol{\mu_{jc}}|\boldsymbol{\bar{\mu}_{j}},\Sigma_{j}(\bar{\kappa}_{j}^{-1}+\kappa_{1}^{-1}))
\end{equation}

In step 4, we derive the posterior of the current class mean vector $\boldsymbol{\mu_{jc}}$ by combining current class sample mean $\boldsymbol{\bar{x}_{jc}}$ from step 1 and the local prior from step 3. 
\begin{gather}
P(\boldsymbol{\mu_{jc}}| \boldsymbol{\mu_{0}}, \Sigma_{j}  ,\kappa_{0}, \kappa_{1},\{\boldsymbol{\bar{x}_{ji}}\}_{t_i=j},\boldsymbol{\bar{x}_{jc}}) \nonumber = N(\boldsymbol{\mu_{jc}} |  \frac{n_{jc}\boldsymbol{\bar{x}_{jc}}+\tilde{\kappa}_{j}\boldsymbol{\bar{\mu}_{j}}}{n_{jc}+\tilde{\kappa}_{j}}, 
\Sigma_{j}(\tilde{\kappa}_{j}^{-1}+n_{jc}^{-1})) \\ 
 \tilde{\kappa}_{j} = \frac{(\sum_{i:t_{i}=j}\frac{n_{ji}\kappa_{1}}{(n_{ji}+\kappa_{1})}+\kappa_{0})\kappa_{1}}{\sum_{i:t_{i}=j}\frac{n_{ji}\kappa_{1}}{(n_{ji}+\kappa_{1})}+\kappa_{0}+\kappa_{1}} \label{eq:mu_ks}
\end{gather}

In step 5, we derive the posterior distribution of the covariance matrix $\Sigma_{j}$ by combining the local prior of the covariance matrix $P(\Sigma_{j}|\Sigma_{0},m)$ with the distribution of the scatter matrices of the classes associated with meta-class $j$ $S_{ji}$ and current class $S_{jc}$:
\begin{gather}
P(\Sigma_{j}|\{S_{ji}\}_{t_i=j}, S_{jc}) = IW(\Sigma_{j}|\bar{S}_c, m+\sum_{i:t_{i}=j} (n_{ji}-1)+n_{jc}) \nonumber\\ 
\bar{S_{c}}  =  \Sigma_{0}+\sum_{i:t_{i}=j}S_{ji}+S_{jc}+S_{\mu}, \quad
S_{\mu}  =  \frac{n_{jc}\tilde{\kappa}_{j}}{\tilde{\kappa}_{j}+n_{jc}}(\boldsymbol{\bar{x}_{jc}}-\boldsymbol{\bar{\mu}_{j}})(\boldsymbol{\bar{x}_{jc}}-\boldsymbol{\bar{\mu}_{j}})^{T}\label{eq:iw} 
\end{gather}

In step 6, we derive the posterior predictive distribution by integrating out meta-class mean vector $\boldsymbol{\mu_j}$ and covariance $\Sigma_j$ in the form of a Student-t distribution as follows. 
\begin{gather}
P(\boldsymbol{x}|\{\boldsymbol{\bar{x}_{ji}}, S_{ji}\}_{t_i=j}, \boldsymbol{\bar{x}_{jc}}, S_{jc}, \boldsymbol{\mu_{0}}, \kappa_0, \kappa_1) = T(\boldsymbol{x}|\boldsymbol{\bar{\mu}_{jc}},\bar{\Sigma}_{jc},\bar{v}_{jc}) \nonumber \\
\boldsymbol{\bar{\mu}_{jc}} = \frac{n_{jc}\boldsymbol{\bar{x}_{jc}}+\tilde{\kappa}_{j}\boldsymbol{\bar{\mu}_{j}}}{n_{jc}+\tilde{\kappa}_{j}}, \quad
\bar{v}_{jc} = n_{jc} + \sum_{i:t_{i}=j}(n_{ji}-1)+m-D+1\nonumber\\
\bar{\Sigma}_{jc} = \frac{\Sigma_{0}+\sum_{i:t_{i}=j} S_{ji}+S_{jc}+S_{\mu}}{\frac{(n_{jc}+\tilde{\kappa}_{j})\bar{v}_{jc}}{n_{jc}+\tilde{\kappa}_{j}+1}} 
\label{eq:i2gmm_seen}
\end{gather}
where, $\boldsymbol{\bar{\mu}_j}$, $\tilde{\kappa}_j$ and $S_{\mu}$ are defined as  in  Equation (\ref{eq:mj_pd}), (\ref{eq:mu_ks}) and (\ref{eq:iw}) respectively. The index $c$ in Equation (\ref{eq:i2gmm_seen}) represents the current seen class, whose PPD is being derived. Top $K$ most similar seen classes are identified as the ones with the smallest Euclidean distance to the current class in the attribute space.  If the current class is a seen class, PPD takes the form in Equation (\ref{eq:i2gmm_seen}). When it is an unseen class with no images available in training, the sample statistics of the current class in (\ref{eq:i2gmm_seen}) drops and PPD becomes: 
\begin{gather}
P(\boldsymbol{x}|\{\boldsymbol{\bar{x}_{ji}}, S_{ji}\}_{t_i=j}, \boldsymbol{\mu_{0}}, \kappa_0, \kappa_1) =  T(\boldsymbol{x}|\boldsymbol{\bar{\mu}_{j}},\bar{\Sigma}_{j},\bar{v}_{j})\nonumber\\
\bar{v}_{j} =  \sum_{i:t_{i}=j}(n_{ji}-1)+m-D+1, \quad
\bar{\Sigma}_{j} = \frac{(\Sigma_{0}+\sum_{i:t_{i}=j} S_{ji})(\tilde{\kappa}_{j}+1)}{\tilde{\kappa}_{j} \bar{v}_{j}}\label{eq:i2gmm_unseen}
\end{gather}
where $\boldsymbol{\bar{\mu}_j}$ and $\tilde{\kappa}_j$  are defined as in  Equation (\ref{eq:mj_pd}) and (\ref{eq:mu_ks}), respectively. 
In this setting, a new image is labeled by evaluating PPDs for seen and unseen classes and assigning the image to the class that generates the maximum likelihood.

\subsection{Meta-class Formation} 
Meta-class for each unseen class is formed by finding K most similar seen classes to the current unseen class using $\mathcal{L}^2$
distance between the attribute vectors ($\phi$) of that unseen class and of seen classes. In the case of tie, the least similar class among selected  seen classes ($K^{th}$)  is replaced by the next one until tie is broken. These define a local prior in the PPD of the unseen class. Meta-class formation for a seen class follows the same procedure. We use the $\mathcal{L}^2$ distance between the current seen class attribute and other seen class attributes to find K most similar classes. As we have access to seen class samples, the PPD of the seen class (Equation~\ref{eq:i2gmm_seen}) uses class samples in addition to local and global priors from its meta-class. 
An illustration for the formation of the meta-class associated with an unseen class \textit{blue whale}, from AWA dataset,  is shown in Figure~\ref{fig:process}.
$\phi$(\textit{blue whale}) is compared against $\phi$(\textit{seen}) in the semantic space, \textit{humpback} and \textit{killer whale} are identified as the two closest matches. Using  \textit{humpback} and \textit{killer} \textit{whale} class samples, the meta-class for \textit{blue whale} is formed  as a local prior in the PPD for \textit{blue whale}.

\section{Experiments}

We evaluate the performance of the proposed approach on several benchmark data sets and compare the results with the current state of the art in ZSL. 

\begin{table}[t]
\begin{center}
\begin{tabular}{c c c c c c c} 
 Dataset & \#imgs & Type & \#att & $|Y^{all}|$ & $|Y^s|$ & $|Y^{u}|$ \\ 
 \hline 
 FLO& 8,189 & fine & 102 & 102 & 62 + 20 & 20 \\
 
 SUN & 14,340 & fine & 102 & 717 & 580 + 65 & 72 \\ 

 CUB & 11,788 & fine & 312 & 200 & 100 + 50 & 50 \\

 AWA1 & 30,475 & coarse& 85 & 50 & 27 + 13 & 10 \\

 AWA2 & 37,322 & coarse & 85  & 50 & 27 + 13 & 10\\

 aPY & 15,339 & coarse & 64 & 32 & 15 + 5 & 12 \\
 \hline
 ImageNet & 14M & large & 500 & 21K & 1K & 20K \\
\end{tabular}
\end{center}
\caption{Specifications of all  datasets used in our experiments. $|Y^{all}|$, $|Y^{s}|$, and $|Y^{u}|$ denote the number of classes in all, seen and unseen classes, respectively. To clarify the numbers in last 3 columns, we give an illustration on FLO dataset: FLO has total of 102 classes of which 62 are training, 20 are validation (both seen during training) and 20 are test classes (unseen during training). }
\label{tab:dataset_summary}
\end{table}

\textbf{Datasets \& specifications.}
Experiments are evaluated on ZSL datasets widely used for benchmarking. Among those, CUB~ \cite{cub}, FLO~\cite{flo} and SUN~\cite{sun} are medium scale, fine-grained datasets. AWA1~\cite{dap_iap} and AWA2~\cite{gbu}  and aPY~\cite{farhadi}, on the other hand, are coarse-grained datasets. Finally we evaluate our model on ImageNet~\cite{imagenet} with more than 14 million images and 21K classes. SUN, AWA1, AWA2, aPY and CUB datasets come with visual attributes whereas FLO uses sentences and ImageNet uses word embeddings as class vectors. We use the publicly available image embeddings of \cite{gbu}, i.e. 2048-dimensional top-layer pooling units of the 101-layered ResNet~\cite{resnet} as feature vectors. Additional information about each dataset including the number of images, number of attributes, and sizes of train, validation, and test class splits are present in Table \ref{tab:dataset_summary}.

For ImageNet following the benchmark in \cite{gbu} we use all of the images from 1K classes, i.e. seen classes, for training so that we do not violate the zero-shot assumption as ResNet-101~\cite{resnet} is trained on the same 1K classes from ImageNet. We evaluate the proposed technique in nine different configurations as proposed in~\cite{gbu}, all of which differs according to how test class subsets are chosen. 

\textbf{Evaluation  criteria.}
We use the same evaluation procedure employed in \cite{gbu} as described below. The standard practice in ZSL literature is to evaluate classification performance by Top-1 accuracy. To avoid large classes dominating the overall accuracy, Top-1 accuracy is separately calculated for each class and the mean of individual class accuracies is used for evaluation. GZSL setting includes both seen and unseen classes in the test phase, hence the search space includes all the classes, i.e. $|Y^{all}|$. Hence, first seen and unseen class accuracies are separately computed and then their harmonic mean is used as the final score for evaluation. For ImageNet, the final score is the average Top-1 accuracy over the images of unseen classes (although the search space is still $|Y^{all}|$) as no images from seen classes are available during testing phase.

\textbf{Implementation details.}
We implement two versions of our model: \textit{unconstrained} (UBZSL) and \textit{constrained} (CBZSL) Bayesian ZSL. For large data sets, e.g. ImageNet, our model in Eq.\ref{eq:i2gmm} suffers from the large memory requirement due to the unconstrained structure of the class covariance matrices. To alleviate this problem we developed a scalable version of our model where the covariance matrices are constrained to have diagonal forms. The only difference between these two models is that constrained version uses an Inverse Gamma prior on the diagonal entries of the covariance matrix as opposed to an Inverse Wishart in the unconstrained version. With this revision the generative model in Eq.\ref{eq:i2gmm} is updated as follows. 
\begin{gather}
\resizebox{\linewidth}{!}{%
$\boldsymbol{x_{jik}^{d}}  \sim  N(\boldsymbol{\mu_{ji}^{d}},\Sigma_{j}^{d})\nonumber, \quad
\boldsymbol{\mu_{ji}^{d}}  \sim  N(\boldsymbol{\boldsymbol{\mu_{j}^{d}}},\Sigma_{j}^{d}\kappa_{1}^{-1})\nonumber \nonumber, \quad
\boldsymbol{\mu_{j}^{d}} \sim N(\boldsymbol{\mu_{0}^{d}},\Sigma_{j}^{d}\kappa_{0}^{-1})\nonumber, \quad
\Sigma_{j}^{d}\sim IG(a_0, b_0)$
}
\label{eq:uc_i2gmm}
\end{gather}
where the superscript $d$ is added to refer to the $d^{th}$ component of each parameter. The Inverse Wishart parameters $m$ and $\Sigma_0$ are replaced with the scale ($a_0$)  and shape ($b_0$) parameters of the Inverse Gamma distribution. The derivation of PPD for the constrained model is in the supplementary. 

The hyperparameters of the model are coarsely tuned to  maximize the harmonic mean score on the validation set for all datasets but ImageNet. The training, test and validation set splits for these datasets are done according to \cite{gbu} to maintain a fair comparison. As hyperparameter tuning for ImageNet can be computationally unmanageable and to demonstrate the robustness of the model we used the hyperparameters of the SUN dataset for ImageNet. For CBZSL  we utilize all 2048 ResNet features whereas for UBZSL we applied PCA to reduce the dimensionality to 500.

Both UBZSL and CBZSL have four hyperparameters: $\kappa_0, \kappa_1, m, s, K$. Here, K is the selected number of classes most similar to the current class in the attribute space. To simplify the parameter tuning process, we set prior mean, $\mu_0$, to the average of class means. We set $\Sigma_0$ to the average of class scatter matrices scaled by a constant $s$. 
Our implementation will be made publicly available upon the completion of the review process. 

\subsection{Model Evaluation}

\begin{wraptable}{r}{0.5\textwidth}
\vspace{-35pt}
    \begin{center}
    \resizebox{\linewidth}{!}{%
    \begin{tabular}{l|c c c c c c}
         \textbf{Method} & \textbf{SUN} & \textbf{CUB} & \textbf{AWA1} & \textbf{AWA2} & \textbf{aPY} & \textbf{FLO} \\ \hline
    UBZSL (V1) & $32.5$ & $24.9$ & $21.1$ & $29.0$ & $10.0$ & $20.5$\\
    UBZSL (V2) & $3.0$ & $18.3$ & $38.0$ & $40.3$ & $9.5$ & $34.1$\\ \hline
    UBZSL  & $32.8$ & $37.5$ & $49.6$ & $49.7$ & $35.4$ & $40.4$
    \end{tabular}
    }
    \end{center}
    \caption{Ablation study (in harmonic mean) on 6 datasets. In the UBZSL (V1) we discard Bayesian aspect and in UBZSL (V2) we impose similar dispersion for meta and actual classes}
    \label{tab:ablation_exp}
    \vspace{-10pt}
\end{wraptable}

\begin{figure}[t]
\begin{centering}
\subfigure[FLO]{\includegraphics[width=0.45\textwidth, trim=0 0 10 0, clip]{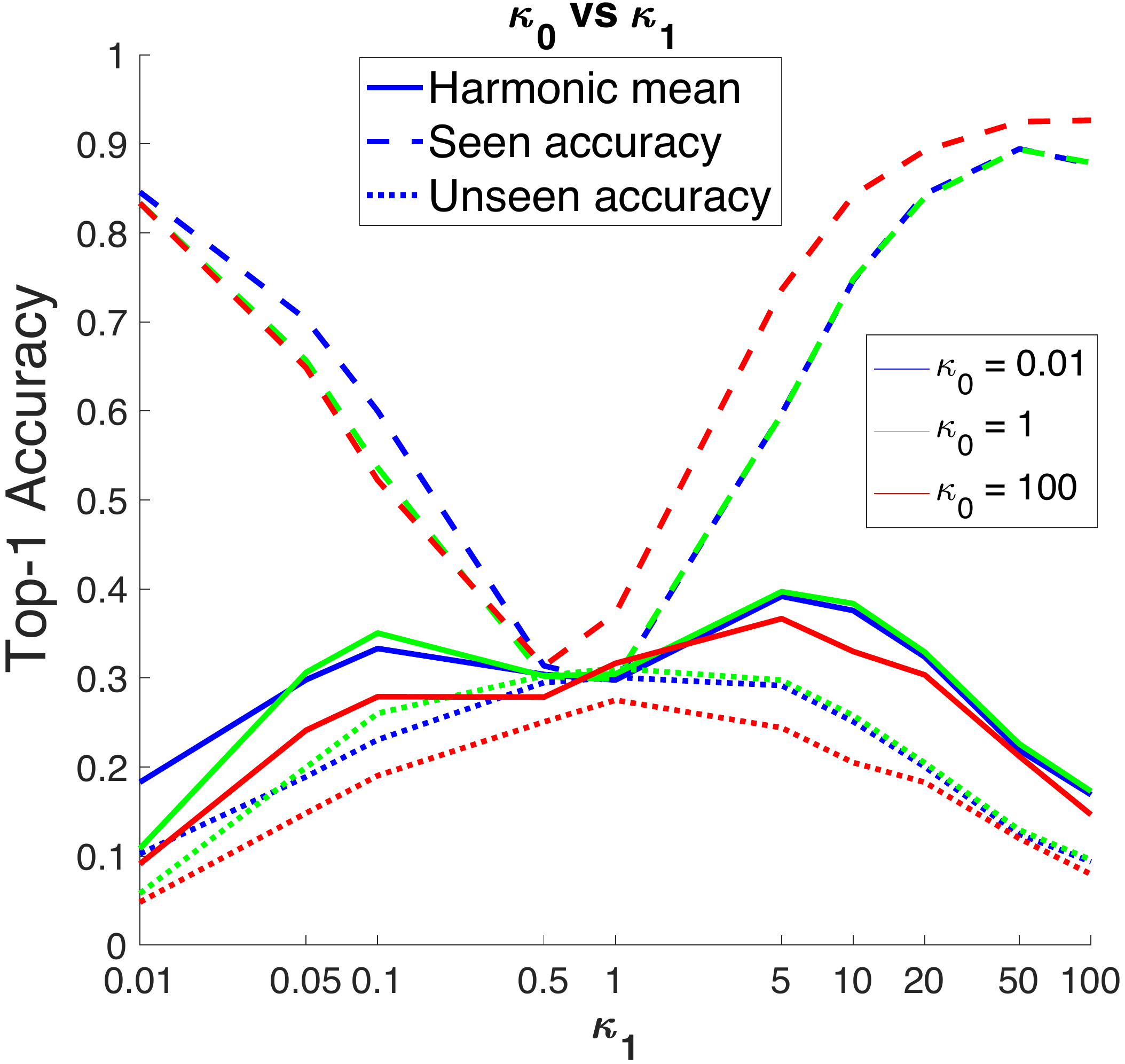}}
\subfigure[AWA2]{\includegraphics[width=0.45\textwidth, trim=0 0 10 0, clip]{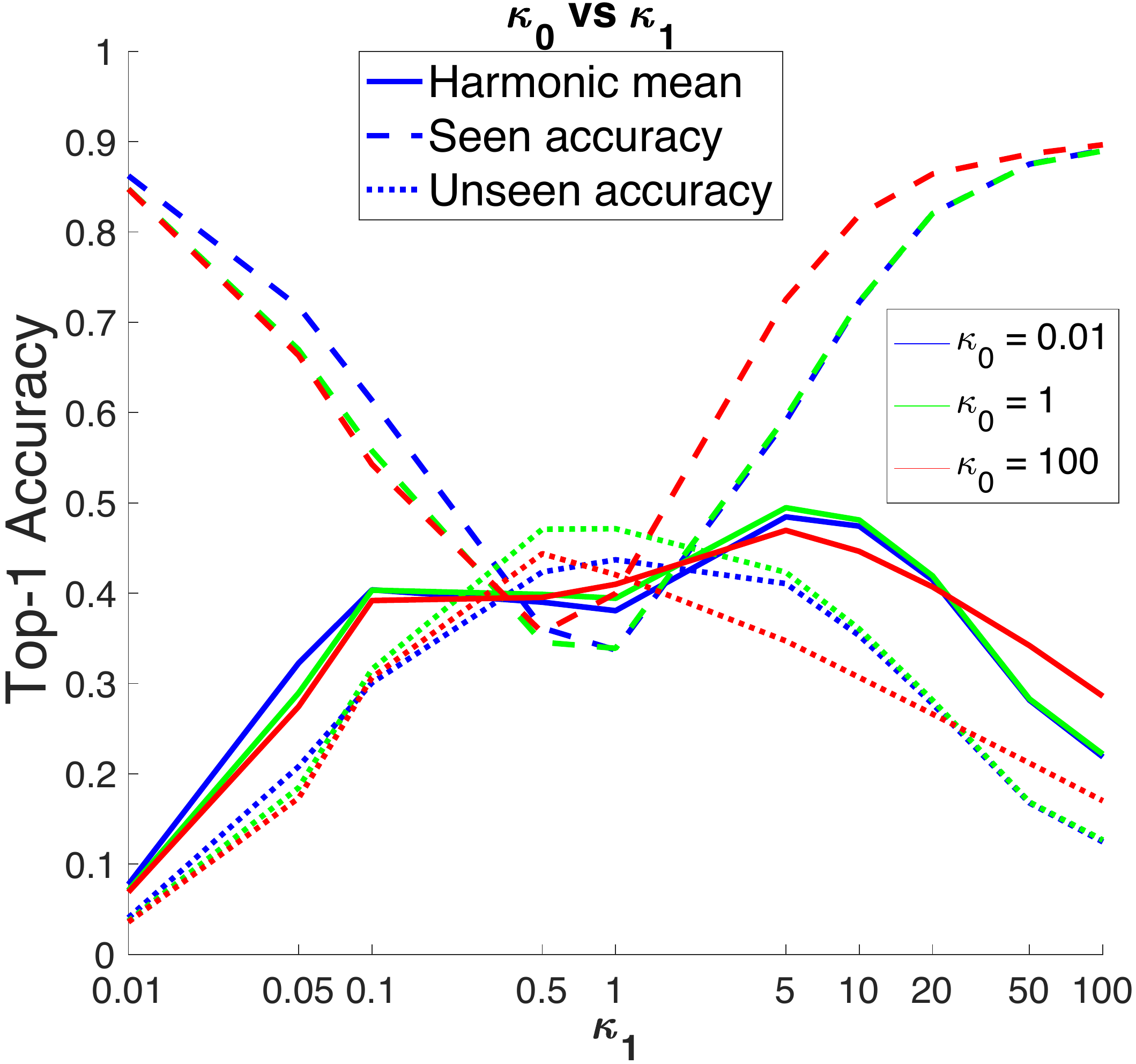}}
\par\end{centering}
\caption{Variations in seen and unseen class accuracies and their harmonic means  with respect to changes in $\kappa_0$ and $\kappa_1$. Seen and unseen class accuracies are highly sensitive to changes in $\kappa_1$ whereas minimal changes are observed wrt changes in $\kappa_0$.}
\label{fig:kappa}
\end{figure}

{
\setlength{\tabcolsep}{3pt}
\renewcommand{\arraystretch}{1.2} 
\begin{table*}[t]
\begin{center}
 \resizebox{\linewidth}{!}{%
\begin{tabular}{l c c c|c c c|c c c|c c c|c c c|c c c}
 & \multicolumn{3}{c}{SUN} & \multicolumn{3}{c}{CUB} & \multicolumn{3}{c}{AWA1} & \multicolumn{3}{c}{AWA2} & \multicolumn{3}{c}{aPY} & \multicolumn{3}{c}{FLO} \\ 
 \textbf{Method} & \textbf{ts} & \textbf{tr} & \textbf{H} & \textbf{ts} & \textbf{tr} & \textbf{H} & \textbf{ts} & \textbf{tr} & \textbf{H} & \textbf{ts} & \textbf{tr} & \textbf{H} & \textbf{ts} & \textbf{tr} & \textbf{H} & \textbf{ts} & \textbf{tr} & \textbf{H} \\ \hline
LATEM\cite{latem} & $14.7$ & $28.8$ & $19.5$ & $15.2$ & $57.3$ & $24.0$ & $7.3$ & $71.7$ & $13.3$ & $11.5$ & $77.3$ & $20.0$ & $0.1$ & $73.0$ & $0.2$ & $6.6$ & $47.6$ & $11.5$ \\
ALE\cite{ale} & $21.8$ & $33.1$ & $26.3$ & $23.7$ & $62.8$ & $34.4$ & $16.8$ & $76.1$ & $27.5$ & $14.0$ & $81.8$ & $23.9$ & $4.6$ & $73.7$ & $8.7$ & $13.3$ & $61.6$ & $21.9$ \\
DEVISE\cite{devise} & $16.9$ & $27.4$ & $20.9$ & $23.8$ & $53.0$ & $32.8$ & $13.4$ & $68.7$ & $22.4$ & $17.1$ & $74.7$ & $27.8$ & $4.9$ & $76.9$ & $9.2$ & $9.9$ & $44.2$ & $16.2$ \\
SJE\cite{sje} & $14.7$ & $30.5$ & $19.8$ & $23.5$ & $59.2$ & $33.6$ & $11.3$ & $74.6$ & $19.6$ & $8.0$ & $73.9$ & $14.4$ & $3.7$ & $55.7$ & $6.9$ & $13.9$ & $47.6$ & $21.5$ \\
ESZSL\cite{eszsl} & $11.0$ & $27.9$ & $15.8$ & $12.6$ & $63.8$ & $21.0$ & $6.6$ & $75.6$ & $12.1$ & $5.9$ & $77.8$ & $11.0$ & $2.4$ & $70.1$ & $4.6$ & $11.4$ & $56.8$ & $19.0$  \\
SYNC\cite{sync} & $7.9$ & $43.3$ & $13.4$ & $11.5$ & $70.9$ & $19.8$ & $8.9$ & $87.3$ & $16.2$ & $10.0$ & $90.5$ & $18.0$ & $7.4$ & $66.3$ & $13.3$ & $-$ & $-$ & $-$ \\
SAE\cite{sae} & $8.8$ & $18.0$ & $11.8$ & $7.8$ & $54.0$ & $13.6$ & $1.8$ & $77.1$ & $3.5$ & $1.1$ & $82.2$ & $2.2$ & $0.4$ & $80.9$ & $0.9$ & $-$ & $-$ & $-$ \\
GFZSL\cite{gfzsl} & $0.0$ & $39.6$ & $0.0$ & $0.0$ & $45.7$ & $0.0$ & $1.8$ & $80.3$ & $3.5$ & $2.5$ & $80.1$ & $4.8$ & $0.0$ & $83.3$ & $0.0$ & $-$ & $-$ & $-$ \\
TCN\cite{tcn} & $31.2$ & $37.3$ & $\textbf{34.0}$ & $52.6$ & $52.0$ & $\textbf{52.3}$ & $49.4$ & $76.5$ & $\textbf{60.0}$ & $61.2$ & $65.8$ & $\textbf{63.4}$ & $24.1$ & $64.0$ & $35.1$ & $-$ & $-$ & $-$ \\

DCN\cite{dcn} & $25.5$ & $37.0$ & $30.2$ & $28.4$ & $60.7$ & $38.7$ & $25.5$ & $84.2$ & $39.1$ & $-$ & $-$ & $-$ & $14.2$ & $75.0$ & $23.9$ & $-$ & $-$ & $-$ \\

REL. NET\cite{Rel_net}\footnotemark[1] & $11.1$ & $20.0$ & $14.3$ & $14.0$ & $35.7$ & $20.1$ & $22.9$ & $76.9$ & $35.3$ & $18.6$ & $87.3$ & $30.6$ & $11.5$ & $60.9$ & $19.4$ & $13.8$ & $73.8$ & $23.2$ \\
\hline
CBZSL & $29.0$ & $32.7$ & $30.7$ & $21.1$ & $43.5$ & $28.5$ & $38.9$ & $67.2$ & $49.3$ & $34.1$ & $72.5$ & $46.4$ & $18.8$ & $70.8$ & $29.6$ & $31.3$ & $28.5$ &	$29.8$\\
UBZSL & $31.7$ & $34.0$ & $32.8$ & $31.5$ & $46.3$ & $37.5$ & $38.7$ & $69.3$ & $49.6$ & $37.1$ & $75.1$ & $49.7$ & $24.0$ & $67.4$ & $\mathbf{35.4}$ & $27.2$ & $78.2$ & $\mathbf{40.4}$
\end{tabular}
}
\end{center}
\caption{GZSL results achieved by the proposed approach (CBZSL and UBZSL) along with results of several other techniques from the literature on SUN, CUB, FLO, AWA1, AWA2, aPY datasets. We measure per-class averages top-1 accuracy on seen classes (\textbf{tr}), unseen classes (\textbf{ts}) and their harmonic mean (\textbf{H}). }
\label{tab:gzsl_results}
\end{table*}
}
In this section, we evaluate our model through an ablation study and investigate the tradeoff between seen and unseen class accuracies. 

\textbf{Model ablation.}
Our model formulates zero-shot learning in the framework of hierarchical Bayes. Towards this end, we validate the necessity of  each component in our model by eliminating one component at a time and investigating the performance of the model with remaining components on several benchmark datasets. 

Our observations from Table \ref{tab:ablation_exp} are as follows.
(1) If we break the hierarchy by removing the meta-class layer, then actual classes are directly linked to the global prior and same PPD is assigned to all unseen classes. Thus, unseen classes can no longer be distinguished during test time. (2) If we discard the Bayesian aspect by eliminating the global and local priors, each seen class is fit a single Gaussian and each unseen class is fit a GMM with K components. We observe drastic drop in harmonic mean, almost cut in half, in all datasets but SUN ($1^{st}$ row in Table \ref{tab:ablation_exp}: V1). In general, GMM works better on fine grained datasets than coarse grained ones as the distribution produced by a mixture of very similar classes can be better fit by GMM compared to a distribution produced by a mixture of relatively less similar classes. (3) Finally,  if we impose similar dispersion for actual and meta classes (by improperly adjusting $\kappa_{0}$ and $\kappa_{1}$) with respect to the center of the data,  harmonic mean again suffers significantly ($2^{nd}$ row in Table \ref{tab:ablation_exp}: V2). In particular, results of the SUN dataset suffers the most.  The impact of improper tuning of $\kappa_1$ is explained in the next section. 
Unlike V1, UBZSL V2 works better on coarse grained datasets (AWA1, AWA2) as class centers in these datasets are more separated than fine grained ones. As a result the adverse effects of setting $\kappa_1<<1$ in experiments performed with these datasets seem to be less significant. 

\footnotetext[1]{As \cite{Rel_net} uses different set of attributes in their experiments, we rerun their algorithm with the attributes from \cite{gbu_tpami} to maintain a fair comparison.}
\textbf{Effect of $\kappa_0$ and $\kappa_1$.}
In both of our models ($\textit{constrained}$ and $\textit{unconstrained}$), different hyperparameter settings can be used to modify the operating point of the classifier to favor seen class accuracy over unseen one or vice versa. In this experiment we investigate the effect of $\kappa_0$ and $\kappa_1$ on seen and unseen class accuracies. Recall that $\kappa_0$ adjusts the dispersion of meta-class centers with respect to the center of the overall data and $\kappa_1$ adjusts the dispersion of actual class centers with respect to their corresponding meta class centers. The smaller these parameters are the higher the dispersion will be. 

Figure \ref{fig:kappa} illustrates on FLO and AWA2 that unseen class accuracy is highest when $\kappa_1$ is close to $1$ and drops significantly lower in both directions, i.e., for  $\kappa_1<<1$ and $\kappa_1>>1$. As expected the opposite of this pattern is observed for seen class accuracy. Although both seen and unseen class accuracies are highly sensitive to the selection of $\kappa_1$, the changes are marginal with respect to $\kappa_0$. Moving $\kappa_1$ towards zero encodes a local prior that imposes unrealistically large dispersion for centers of actual-classes sharing the same meta-class, which violates the main assumption of our model that classes sharing the same meta class are semantically similar classes. On the other hand moving $\kappa_1$ towards infinity encodes a local prior that imposes limited to no deviation among centers of actual classes which is another extreme that is not true for real-world datasets, i.e. classes are supposed to be statistically identifiable. 

In both extremes unrealistic prior assumptions that cannot be reconciled with the characteristics of real-world data sets impede knowledge transfer between seen and unseen classes and lead to poor classification performance on unseen classes. On the other hand, the same extreme assumptions happen to help with seen class accuracies because likelihood and data-driven local priors (both of which lacks for unseen classes) outweigh the effect of unrealistic global prior in posterior predictive distributions. 

\subsection{Comparison with State of the Art}
Results obtained by the proposed CBZSL and UBZSL models on SUN, CUB, FLO, AWA1, AWA2, aPY datasets are presented in Table \ref{tab:gzsl_results}. In addition to all SotA techniques 
reported in \cite{gbu_tpami} we also included results of more recently published techniques \cite{Rel_net,dcn,tcn} in this comparison. These results suggest that the proposed unconstrained model (UBZSL) demonstrates better performance than all other techniques but TCN. The constrained version of our model, i.e., CBZSL, also renders comparable results with the unconstrained version of the model despite its simplicity. 

Results in Table \ref{tab:gzsl_results} further show that in all of the experiments, unseen class accuracies achieved by our models are substantially higher than those achieved by all other techniques, but the TCN~\cite{tcn} model. This is achieved while maintaining a comparable performance on seen class accuracies in most of the experiments. Intuitively speaking, the two-level Bayesian hierarchy defined by meta-classes is expected to better manage the open space risk \cite{openset} by assigning an image of an unseen class to its meta class as opposed to misclassifying it into one of the seen classes. 

\begin{wraptable}{r}{0.5\textwidth}
\vspace{-25pt}
{
\setlength{\tabcolsep}{3pt}
\renewcommand{\arraystretch}{1.1} 
\centering
\resizebox{\linewidth}{!}{%
\begin{tabular}{l r r r| r r r| r r r}

 & \multicolumn{3}{c}{\textbf{UBZSL}} & \multicolumn{3}{c}{\textbf{CBZSL}}  & \multicolumn{3}{c}{\textbf{SoA from} \cite{gbu_tpami}} \\ 
 \textbf{Split} & \textbf{1} & \textbf{5} & \textbf{10} & \textbf{1} & \textbf{5} & \textbf{10} & \textbf{1} & \textbf{5} & \textbf{10}   \\ \hline
2Hop & $2.6$ & $13.1$ & $20.3$ & $\mathbf{3.9}$ & $\mathbf{15.0}$ & $\mathbf{22.8} $ & $2.2$ & $10.3$ & $19.3$\\
3Hop & $0.8$ & $4.1$ & $6.9$ & $\mathbf{1.0}$ & $\mathbf{4.1}$ & $6.9$ & $0.8$ & $3.7$ & $\mathbf{7.2}$\\
Lp500 & $1.8$ & $5.1$ & $8.6$ & $\mathbf{2.5}$ & $\mathbf{10.2}$ & $\mathbf{14.3}$ & $1.9$ & $6.1$ & $10.4$\\
Lp1K & $1.2$ & $4.6$ & $7.3$ & $\mathbf{2.3}$ & $\mathbf{7.3}$ & $\mathbf{10.7}$ & $1.4$ & $4.8$ & $8.5$\\
Lp5K & $0.5$ & $2.0$ & $3.5$ & $\mathbf{0.6}$ & $\mathbf{2.4}$ & $\mathbf{4.0} $ & $0.4$ & $2.2$ & $3.9$\\
Mp500 & $3.4$ & $17.4$ & $26.5$ & $\mathbf{7.5}$ & $\mathbf{25.2}$ & $\mathbf{35.0}$ & $2.9$ & $14.9$ & $26.6$ \\
Mp1K & $2.4$ & $13.0$ & $20.2$ & $\mathbf{4.8}$ & $\mathbf{17.3}$ & $\mathbf{25.5}$ & $2.3$ & $11.8$ & $20.7$\\
Mp5K & $1.1$ & $6.1$ & $9.9$ & $\mathbf{1.5}$ & $\mathbf{6.6}$ & $\mathbf{10.5}$ & $1.1$ & $6.2$ & $10.0$\\
All & $0.3$ & $1.8$ & $3.0$ & $\mathbf{0.4}$ & $1.8$ & $2.9$ & $0.3$ & $\mathbf{2.0}$ & $\mathbf{3.4}$\\
\end{tabular}
}
\caption{ImageNet results in nine different test phase configurations. Lp and Mp refer to least and most populated classes, respectively. 2/3 Hop represents the classes that are 2/3-hops away from 1K training classes according to the ImageNet label hierarchy. Finally All appears for all 21K ImageNet classes. The results are in  top-K accuracy.}
\label{tab:gzsl_in}
\vspace{-0.4cm}
}
\end{wraptable}

\subsection{Large-Scale Experiments on ImageNet}
ImageNet is currently the most challenging dataset for ZSL. Arguably it constitutes the most natural setup to evaluate ZSL learning performance as it contains 22K classes (1K of which are used to train state of the art deep neural networks) and most of these classes are sparsely populated.

Table \ref{tab:gzsl_in} summarizes ImageNet results under nine different test set configurations. Our unconstrained model (UBZSL) improves over the state of the art in 2/3 Hop and highly populated test classes. Of particular importance is the highly competitive performance by the constrained model (CBZSL) that improves the current state of the art in all test configurations with respect to Top-1 accuracy (3.9\% vs 2.18\% on 2Hop, 7.51\% vs 2.9\%  on Mp500, 4.78\% vs 2.34\% on Mp1000). Our model achieves the best results in eight of the nine test configurations for Top-5 and seven of the nine for Top-10 accuracies. Especially in the least populated (Lp500) classes the accuracy improvement is four percentage points in Top-5 and Top-10 accuracies. In most populated classes (Mp500) the accuracy gets almost doubled, i.e. $25.20\%$ vs $14.86\%$ on Top-5. 

These results show that as the number of classes and the average number of samples per class ($1300$ in ImageNet vs $700$ in benchmark datasets)  increase, the explicit hierarchy across classes becomes more evident leading to more informative local priors. 
ImageNet contains both coarse- and fine-grained classes. The results suggest that our technique can be equally effective on datasets with hybrid granularity.

\section{Conclusions}

\noindent \textbf{Summary of our contributions:}  In this study, we proposed a Bayesian approach to ZSL that relies on the consideration that classes in real-world datasets emerge at different levels of abstraction, and there are meta-classes that inherently organize the class hierarchy in the semantic space. We introduced concepts of local and global priors and showed that knowledge transfer from seen classes to unseen ones could be effectively carried out in the image space by a two-layer GMM. The proposed two-layer GMM offers extreme flexibility in modeling datasets with different characteristics by tuning its hyperparameters, each of which models a different aspect of the data. We performed extensive experiments with benchmark datasets (fine-grained, coarse-grained, and large-scale) to demonstrate the utility of the proposed Bayesian approach for ZSL, which favors the proposed approach over other state-of-the-art inductive ZSL techniques. 

\bigskip
\noindent \textbf{Future Research Directions:}  Recently proposed transductive methods \cite{f_clsgan,cada_vae,vae_visualization} have proved that generating features for unseen classes and treating ZSL as a closed-set classification can produce much better results than running ZSL in an inductive setting. Although features generated by these techniques do not seem to preserve correlation among features and are far from recovering unseen class distributions, they do preserve the relative distances among unseen classes, which in turn helps improve the performance of a softmax classifier in the closed-set setting. Thus, using prototypical feature vectors for unseen classes and integrating these into PPDs can offer significant boost for the performance of the proposed hierarchical Bayesian model. In our future work we aim to demonstrate that these prototypical feature vectors can be easily obtained by solving a simple compressed sensing problem and PPDs updated with these prototypical vectors  can be used to generate new features in a probabilistic way. Such an approach can potentially preserve both the correlation among features and the relative distance between classes to generate more realistic features. Although not discussed in current work the proposed framework can be easily and effectively extended for any-shot learning problems, which will be a research direction we will pursue in parallel to probabilistic feature generation.

\bigskip
\noindent \textbf{Acknowledgements.} This work has been partially funded by the NSF grant 1252648 - ISS (M. D.), 
the ERC grant 853489 - DEXIM (Z. A.) and by DFG grant under Germany’s Excellence Strategy – EXC number 2064/1 – Project number 390727645.
\pagebreak

\bibliographystyle{splncs04}
\bibliography{egbib}

\end{document}